\documentclass[journal,12pt,onecolumn,draftclsnofoot]{IEEEtran}
\usepackage{xcolor,soul,framed} 
\usepackage{setspace}

\colorlet{shadecolor}{yellow}
\usepackage[pdftex]{graphicx}
\graphicspath{{../pdf/}{../jpeg/}}
\DeclareGraphicsExtensions{.pdf,.jpeg,.png}

\usepackage[cmex10]{amsmath}
\usepackage{array}
\usepackage{mdwmath}
\usepackage{mdwtab}
\usepackage{eqparbox}
\usepackage{url}
\usepackage{algorithm}
\usepackage{algpseudocode}
\usepackage[percent]{overpic}
\usepackage{amsmath}
\usepackage{amsfonts}
\usepackage{authblk}
\usepackage{hyperref}

\begin{document}

\title{Deep Learning Enabled Time-Lapse 3D Cell Analysis}
\author{Jiaxiang Jiang$^{\star}$$^{\ast}$, Amil Khan$^{\star}$, S. Shailja$^{\star}$, Samuel A. Belteton$^{\dagger}$, Michael Goebel$^{\star}$, Daniel B. Szymanski$^{\dagger}$,  and B.S. Manjunath$^{\star}$$^{\ast}$}
\affil{$^{\star}$ Department of Electrical and Computer Engineering, University of California, Santa Barbara \\$^{\dagger}$ Department of Botany and Plant Pathology, Purdue University}


\maketitle

\begin{abstract}

This paper presents a method for time-lapse 3D cell analysis. Specifically, we consider the problem of accurately localizing and quantitatively analyzing sub-cellular features, and for tracking individual cells from time-lapse 3D confocal cell image stacks. 
The heterogeneity of cells and the volume of multi-dimensional images presents a major challenge for fully automated analysis of morphogenesis and development of cells. 
This paper is motivated by the pavement cell growth process, and building a quantitative morphogenesis model.
We propose a deep feature based segmentation method to accurately detect and label each cell region. 
An adjacency graph based method is used to extract sub-cellular features of the segmented cells.  
Finally, the robust graph based tracking algorithm using multiple cell features is proposed for associating cells at different time instances.
Extensive experiment results are provided and demonstrate the robustness of the proposed method. 
The code is available on Github \footnote{\url{https://github.com/UCSB-VRL/Time-lapse3DCellAnalysis}} and the method is available as a service through the BisQue portal.
\footnote{$^{\ast}$ corresponding authors}
\end{abstract}


%


\section{Introduction}


The sizes and shapes of leaves are key determinants of the efficiency of light capture in plants, and the overall photosynthetic rates of the canopy is a key determinant of yields \cite{leafimportanceinag}. The rates and patterns of leaf expansion are governed by the epidermal tissue \cite{epidermisimportance} but understanding how the irreversible growth properties of its constituent jig-saw-puzzle piece cells related to organ level shape change remains as a major challenge.

\begin{figure*}[ht]
    \centering
    \includegraphics[width=1.0\textwidth ]{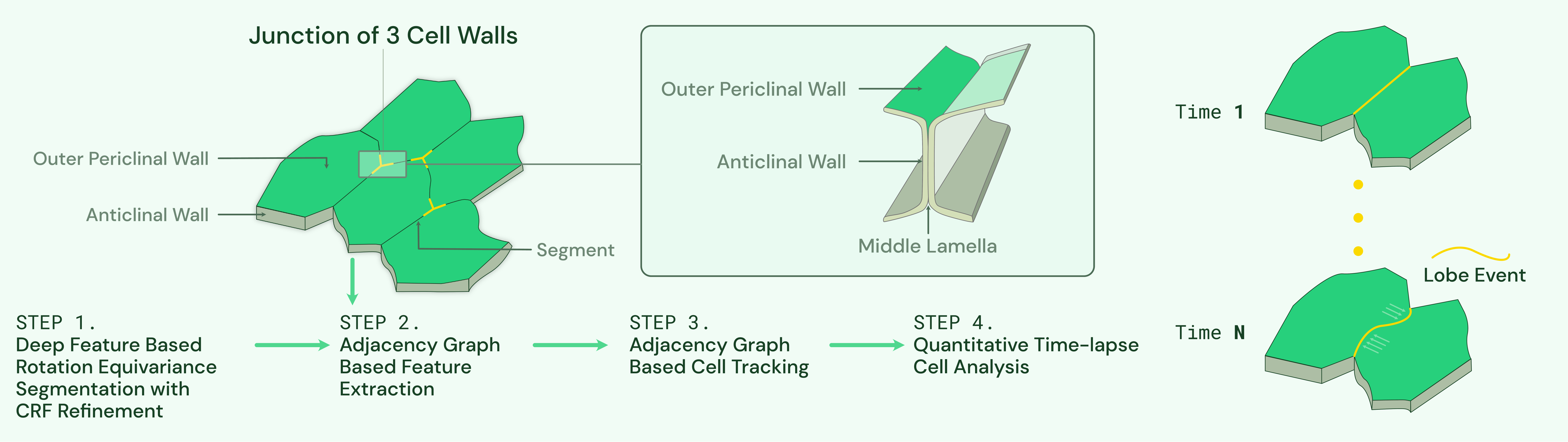}
    \caption{Workflow of proposed method. Modified from \cite{belteton2021real}. Given a sequence of 3D image stacks, deep feature based rotation equivariance deep learning model with CRF refinement is used to segment each cell. Then adjacency graph is built based on segmented image and used for sub-cellular feature extraction and tracking. Sub-cellular features such as junction of three cell walls and anticlinal wall segment are illustrated in the figure. Next detected segments will be used in \cite{lobefinder} to detect lobes. This paper mainly focuses on Step 1 to Step 3.}
    \label{fig:workflow}
\end{figure*}

The epidermal cell, also known as pavement cell, undergoes a dramatic transformation from a slightly irregular polyhedral cell to a highly convoluted and multi-lobed morphology. The interdigitated growth mode is widespread in the plant kingdom \cite{vHofely2019puzzles}, and the process by which lobing occurs can reveal how force patterns in the tissue are converted into predictable shape change \cite{belteton2021real}. To analyze the slow and irreversible growth behavior across wide spatial scales, it is important to track and map lobing events in the epidermal tissue.
It has been shown that cell walls perpendicular to the leaf surface, the anticlinal wall as illustrated in Fig.~\ref{fig:workflow}, can be used to detect new lobe formations \cite{lobefinder,pacequant}.



Time-lapse image stacks from 3D confocal imagery provide a good resource to study the pavement cell growth process, and build the quantitative cell morphogenesis model \cite{purdue, belteton2021real}. 
3D confocal microscopy data contain large amount of cell shape and sub-cellular cell wall structure information.
Cell analysis requirements include detecting sub-cellular features such as junctions of three cell walls and segments shape of anticlinal cell walls used to detect lobes, all of which depends on accurate segmentation. These sub-cellular features are illustrated in Fig.~\ref{fig:workflow}. 
Currently, these features are usually acquired manually from 3D image stacks. 
Manual extraction and analysis is not only laborious but also prevents evaluation of large amounts of data necessary to map relationships between lobe formation to leaf growth.


Existing automatic time-lapse cell analysis methods include mainly two steps: (1) Recognizing and localizing cells and cell walls spatially (segmentation) and tracking cells in temporal dimension, (2) cellular/sub-cellular feature extraction. Both of which are existing challenges with automated analysis systems.


There is an extensive literature on cell segmentation \cite{cellunet,RACE,mosaliganti2012acme,mars,supervoxelmerge,cellseg2,GANsegmentation,cellect,stringer2021cellpose,nucleicell} and tracking \cite{track1,track2}. In \cite{RACE,mosaliganti2012acme,mars,cellect} morphological operations are first used to denoise the images followed by watershed or level set segmentation methods to get the final cell segmentation. In \cite{nucleicell}, the nuclei information is provided for accurate cell segmentation. However, these methods do not provide accurate localization of the cell wall features with only cell boundary information that are needed for quantification. 
In \cite{cellunet,cellseg2,GANsegmentation}, they focus on improving the cell boundary segmentation accuracy.
In \cite{cellseg2,GANsegmentation}, they treat the cell segmentation problem as a semantic segmentation problem, using Generative Adversarial Networks (GAN) to differentiate between boundary pixels, cell interior, and background. These methods provide respectable accuracy on cell boundaries but they are not guaranteed to give a closed cell surface. 
The method proposed in \cite{stringer2021cellpose} that can provide closed 2D surface while maintaining good 2D cell segmentation boundary results.
It is challenging to do the downstream cell analysis such as cell tracking without a closed 3D cell surface.
Based on segmentation or detection of cells, \cite{track1,track2} rely on Viterbi algorithm to track cells. They require the global optimization which is inefficient to get the cell trajectory.


This paper presents a robust, time-lapse cell analysis method building upon our earlier work \cite{cellunet}. In 
\cite{cellunet} we use Conditional Random Field (CRF) to get the improved cell boundaries while maintaining a closed cell surface. To make the segmentation method more robust to different datasets, we propose a modification to \cite{cellunet} that incorporates rotation invariance in the 3D convolution kernels. A segmentation map labeling each individual cell in the 3D stack is thus created and a cell adjacency graph is constructed from this map.
The adjacency graph is an undirected weighted graph with each vertex representing a cell and the weight on the edge representing the minimum distance between two cells. 
Based on this adjacency graph, sub-cellular features illustrated in Fig.~\ref{fig:workflow} are computed. 
The cells are tracked by comparing the corresponding adjacency graphs in the time sequence similar to our previous work \cite{nucleitrack}.
Details of the complete workflow will be described in section 3.

We demonstrate the performance of the proposed \textit{segmentation} method on multiple cell wall tagged data sets.
The performance of the \textit{tracking} method is demonstrated both on cell wall tagged and nuclei tagged imagery.

In summary, the main contributions of this paper include
\begin{itemize}
    \item The first deep learning enabled end-to-end fully automated time-lapse 3D cell analysis method
    \item A new 3D cell segmentation network with rotation equivariance that is robust to different imaging conditions
    \item A novel graph based method for multiple instance tracking and sub-cellular feature extraction as well as the novel evaluation metrics to evaluate sub-cellular feature extraction accuracy 
    \item We will release a new membrane tagged imagery with partially (expert) annotated sub-cellular features and fully annotated by our computational method.
\end{itemize}

%
%

\section{Method}



Our cell analysis method is illustrated in Fig.~\ref{fig:workflow}. 
First, we segment cells from each image stack in the time sequence. 
Second, the adjacency graph is built based on segmented images and is used to compute sub-cellular features and cell tracking features.
Finally, quantitative measurements of the cell segmentation (cell wall, cell count, cell shape), sub-cellular features (junctions of three cell walls detection accuracy, anticlinal wall segment shape), and tracking results are provided.

\subsection{Segmentation}

\begin{figure}[htb]
  \centering
  \centerline{\includegraphics[width=1.0\columnwidth]{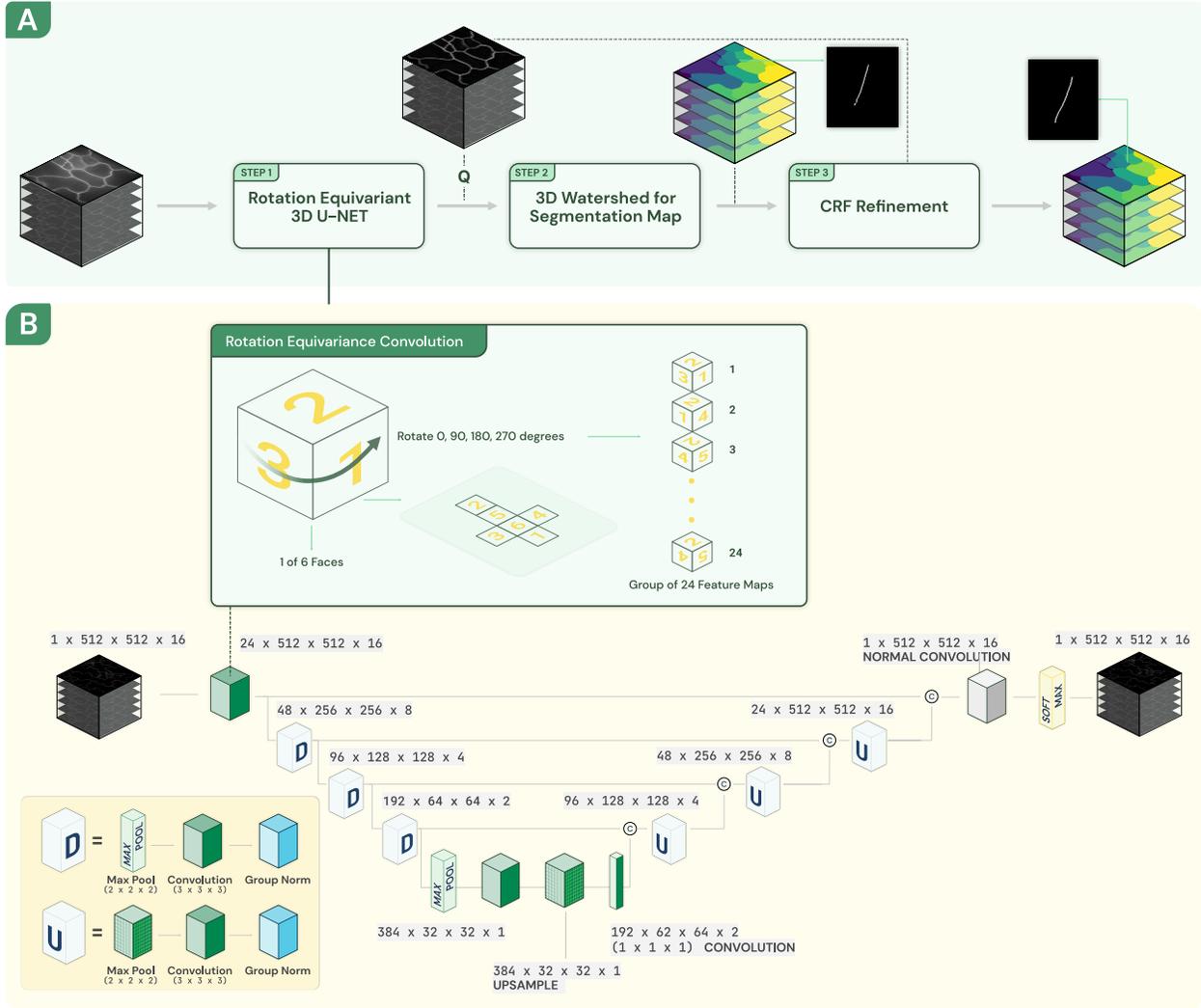}}
  \caption{A. Segmentation workflow includes rotation equivariant 3D U-Net, 3D watershed, and CRF refinement. B. In 3D equivariant U-Net, all convolution layers are rotation equivariant convolution layers. The raw 3D image stack is truncated into 16 slices and then input to 3D equivariant U-Net.}
  \label{fig:segmentaiton_flow}
\end{figure}

We adopt the cell segmentation workflow from \cite{cellunet} with rotation equivariance constrained enforced as shown in Fig.\ref{fig:segmentaiton_flow}.
3D U-Net is a reliable method for semantic segmentation specifically for biomedical images, and 2D rotation equivariance has shown its robustness to input image orientation \cite{rotation_equiv}.
Therefore, we first use a rotation equivariance 3D U-Net to generate a probability map of each voxel being a cell wall.
The full 3D U-Net rotation equivariance is achieved by replacing all convolution layers with rotation-equivariant layers described in the next paragraph. 
Second, to make sure we can get closed cell surfaces, a 3D watershed algorithm whose seeds are generated automatically is applied to the cell wall probability map, and outputs the initial cell segmentation result.  The initial cell segmentation boundary is closed but may not be smooth because watershed segmentation is sensitive to noise. Finally, a conditional random field (CRF) model is used to refine the cell boundaries of the initial cell segmentation. The CRF model takes the cell wall probability map and initial cell segmentation labels as input and outputs a smooth and closed cell wall. 
In the following section, we will discuss the details of our  rotation-equivariant convolution layers and the use CRF to refine the cell segmentation boundary.

3D rotation-equivariant layers are a generalization of convolution layers and are equivariant under general symmetry groups, such as the group of four $90^{\circ}$ 2D rotations \cite{rotation_equiv}. The corresponding 3D rotation group has 24 rotations as illustrated in Fig. \ref{fig:segmentaiton_flow} (A cube has 6 faces and any of those 6 faces can be moved to the bottom, and then this bottom face can be rotated into 4 different positions). To achieve this, convolution operations on feature maps are operating on a group of features which implies that we should have feature channels in groups of 24, corresponding to 24 rotations in the group. 

For a given cell wall probability map \textbf{Q} and cell labels \textbf{X}, the conditional random field is modeled by the Gibbs distribution,
\begin{equation}
P(\textbf{X}|\textbf{Q})=\frac{1}{Z(\textbf{Q})}\exp(-E(\textbf{X}|\textbf{Q}))
\end{equation}
where denominator $Z(\textbf{Q})$ is the normalization factor. 
The exponent is the Gibbs energy function and we need to minimize the energy function $E(\textbf{X})$ to get the final refined label assignments (for notation convenience, all conditioning is omitted from this point for the rest of the paper). 
In the dense CRF model, the energy function is defined as 
\begin{equation} \label{eq:energy}
E(\textbf{X})=\sum_i\psi_u(x_i)+\sum_{i<j}\psi_p(x_i,x_j)    
\end{equation} 
where $i$ and $j$ are the indices of each voxel which iterate over all voxels in the graph, and $x_i$ and $x_j$ are the cell labels of vertices $i$ and $j$.
$i,j\in \{ 1,2,...,N \}$ and $N$ is the total number of voxels in the image stack.
$x_i,x_j\in \{ 0,1,2,...,L \}$ and $L$ is the total number of cells identified by the watershed method (0 is the background class). 
The first term of eq.~\ref{eq:energy}, the unary potential, is used to measure the cost of labeling $i_{th}$ voxel as $x_i$ and it is given by
$\psi_u(x_i)=-\log{P(x_i)},$
where $P(x_i)$ is the probability of voxel $i$ having the label $x_i$. It is initially calculated based on the cell wall probability map \textbf{Q} and the label image of the watershed $\textbf{X}^0$ (The superscript 0 is used to denote the initial cell label assignment after watershed).  $P(x_i^0)=1-q_i$ if voxel $i$ is inside the cell with label $x_i^0$ after the watershed or if $x_i^0$ is the background label, and $P(x_i^0)=0$ otherwise. 
$q_i$ is the $i_{th}$ voxel value in the probability map from the rotation equivariant 3D U-Net. 
$1-q_i$ represents the probability of voxel being the interior point of the cell.
The pairwise potential in eq.~\ref{eq:energy} takes into account the label of neighborhood voxels to make sure the segmentation label is closed and the boundary is smooth \cite{crf}. 
It is given by:
\begin{equation}
    \psi_p(x_i,x_j)=\mu(x_i,x_j)\sum_mw^{(m)}k^{(m)}(\textbf{f}_i,\textbf{f}_j)
\end{equation}
 where the penalty term $\mu(x_i,x_j)=1$ if $x_i\neq x_j$, and $\mu(x_i,x_j)=0$ otherwise. $w^{(m)}$ is the weight for each segmentation label $m\in \{0,1,2,...,L\}$, and $k^{(m)}$ is the pairwise kernel term for each pair of voxels $i$ and $j$ in the image stack regardless of their distance that capture the long-distance voxel dependence in the image stack. 
 $\textbf{f}_i$ and $\textbf{f}_j$ are feature vectors from the probability map \textbf{Q}. 
$\textbf{f}_i$ incorporates location information of voxel $i$ and the corresponding value in the probability map: $\textbf{f}_i = <\textbf{p}_i,q_i>$ where $\textbf{p}_i = <x_i, y_i, z_i>$, and $x_i, y_i$ and $z_i$ are the voxel $i$ in coordinates. Specifically, the kernel $k(\textbf{f}_i,\textbf{f}_j)$ is defined as
 \begin{equation}
     k(\textbf{f}_i,\textbf{f}_j)={\gamma_1\exp(-\frac{||\textbf{p}_i-\textbf{p}_j||^2}{2\sigma^2_\alpha}-\frac{||q_i-q_j||^2}{2\sigma^2_\beta})+\gamma_2\exp(-\frac{||\textbf{p}_i-\textbf{p}_j||^2}{2\sigma^2_\gamma}})
 \end{equation}
where the first term depends on voxel location and the corresponding voxel value in probability map. The second term only depends on the voxel location. $\sigma_\alpha$, $\sigma_\beta$, and $\sigma_\gamma$ are the hyper parameters that depends on shape and size of cells in each dataset. Finally, we pick the best label assignment $\textbf{X}^\ast$ as the final cell segmentation that minimizes energy function $E(\textbf{X})$. The efficient CRF inference algorithm described in \cite{crf} is used to find $\textbf{X}^\ast$ which is the final cell segmentation mask. In our experiments, the above mentioned CRF regularisation is iteratively applied 2-4 times.


\begin{figure*}[ht]
    \centering
    \includegraphics[width=1.0\textwidth ]{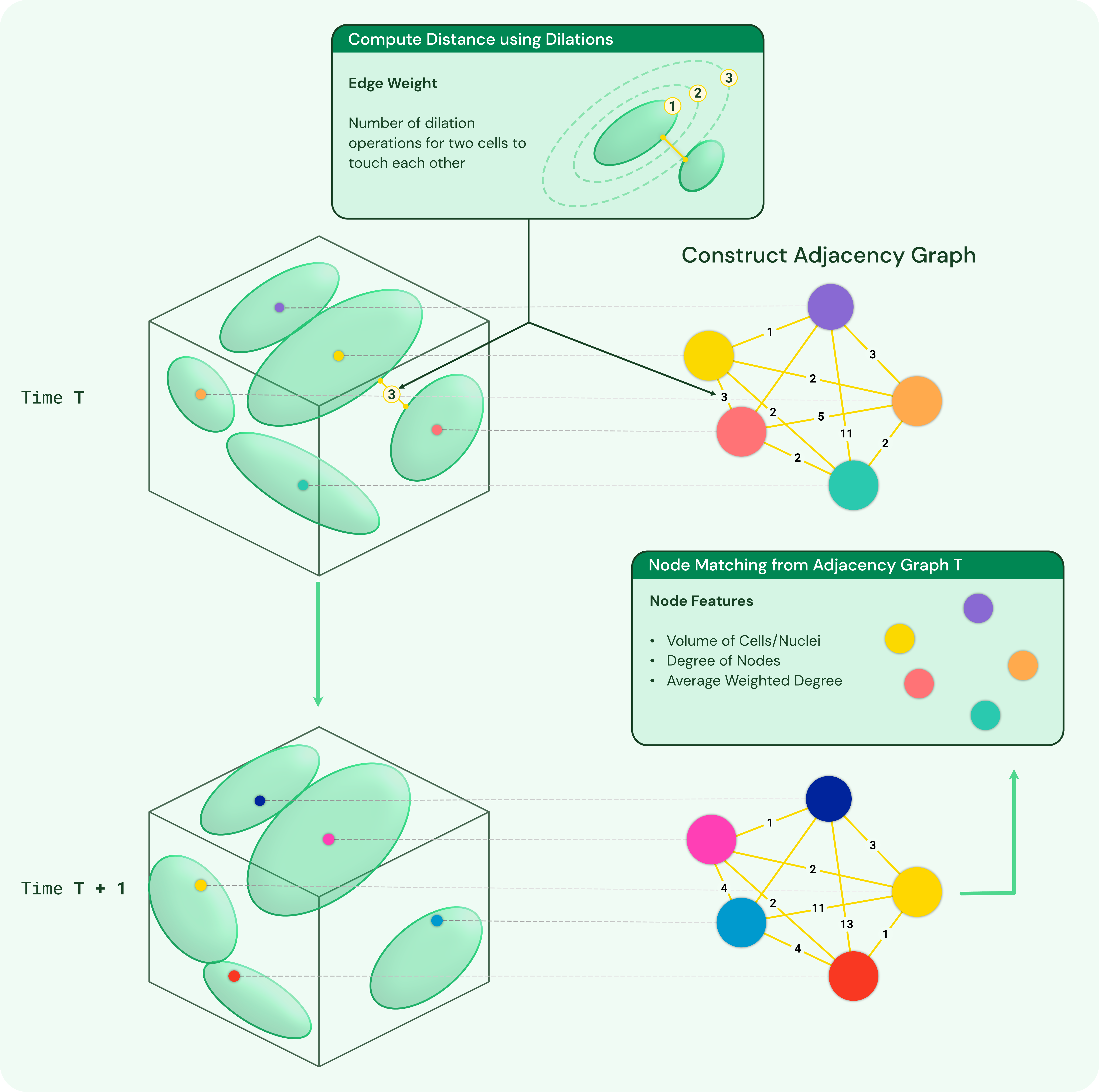}
    \caption{Constructing adjacency graph from the segmentation image and tracking cells/nuclei in consecutive frames using adjacency graph node features. Color of nodes denote the label/track of the cell/nuclei. Initially, random labels are assigned for each node in the adjacency graph. For T+1 frame, after node matching for time T, track IDs are assigned to each node in T+1.}
    \label{fig:tracking}
\end{figure*}

\subsection{Tracking and Feature Computation}
After segmentation of 3D image stacks, the cells are detected and labeled in 3D space. 
Next, we utilize 3D spatial location of cells to build the adjacency graph for sub-cellular feature extraction and tracking as illustrated in Fig.\ref{fig:tracking}.

Adjacency graph $G(V,E)$ is a weighted undirected graph. 
The vertex $v_i \in V$ represents the $i-$th cell. 
For each pair of vertices $(v_i,v_j)$, there is an edge $e_i \in E$ connecting them. 
The weight $w_i \in W$ of the edge $e_i$ is the distance between cell $i$ and $j$. 
The distance between two cells is computed as the number of morphology dilation operations needed of cells $i$ and $j$ until cell $i$ and $j$ become a single connected component. The details of this adjacency graph construction are given in Algorithm 1 below.

  \begin{algorithm}
   \caption{Cell Adjacency Graph Construction}
    \begin{algorithmic}
      \Function{CellAdjacency}{Segmented Image Stack}
        \State Initiate AdjacentGraph
        \State Initiate drawboard (the same shape as segmentation)to be 0
        \For{$i = 1$ to number of cells}
            \For{$j = 1$ to number of cells}
            \State entry of drawboard is set to 1 if the voxel belong to cell i or j
            \If{$i \neq j$}
                \State 3D dilation of drawboard
                \State connected component analysis on drawboard                \If {number of component is 1}
                    \State append j to ith entry of AdjacentGraph
                \EndIf
            \EndIf
            \EndFor
        \EndFor
       \EndFunction

\end{algorithmic}
\end{algorithm}

\subsubsection*{Sub-cellular Feature Extraction Using Adjacency Graph}
Sub-cellular feature extraction is based on the graph representation of the segmented image. 
To compute the anticlinal wall segments of cell $i$, we find all neighbor cells of cell $i$. 
The neighbor cells $\textbf{N}_i$ are defined to be all cells that are at a distance 1 from cell $i$. 
The anticlinal wall segments is found by collecting all points in the segmentation image shared by cell $i$ and cell $j$ where cell $j \in \textbf{N}_i$.
To compute the junctions of 3 cell walls, we first pick cell $j \in \textbf{N}_i$. Then the junctions of 3 cell walls is computed as the points in the segmentation image shared by cell $i$, cell $j$, and cell $k$ where cell $k$ is $\textbf{N}_i \cap \textbf{N}_j$.

\subsubsection*{Tracking Using Adjacency Graph}
The assumption we make for the cell tracking is that in consecutive image stacks, cells should have similar relative location. 
For this, we will focus on computing features $\mathbf{f}_{loc}$ that represent cell relative location information 
derived from the adjacency graph.

Cell location feature vector $\mathbf{f}_{loc}$ is a two dimensional vector $(N,D)$, where $N$ is the total number of neighbor cells and $D$ is the average distance from all other cells. 
Consider the adjacency graph $G(V,E)$ of the segmented image stack. 
For node $i$ in the graph, the location feature vector can be expressed as:  
\begin{equation}
    \mathbf{f}_{loc}^i=(N,D)=(deg(v_i),\text{wdeg}(v_i)) 
\end{equation}
where $v_i \in V$, $deg(v_i)$ is the cardinality of $N_i$, and $\text{wdeg}(v_i)$ is the weighted degree of the vertex $v_i$. The weighted degree of the vertex $v_i$ is defined as:
\begin{equation}
    \text{wdeg}(v_i)=\frac{\sum_{j} w_{ij}}{\text{degree}(v_i)}
\end{equation}
where $\text{degree}(v_i)$ represents the degree of the vertex $v_i$. 
Then we compute the cell size by counting number of voxels within the cell.
Combining the cell location feature and cell size feature, we get the three dimensional feature vector $\mathbf{f}_{\text{track}}$. 
The details of algorithm used for calculating $\mathbf{f}_{\text{track}}$ is described below:
 \begin{algorithm}
   \caption{Cell Tracking Feature Computation}
    \begin{algorithmic}
      \Function{CellTrackFeature}{Segmentation and G(V,E)}
        \State Initiate FeatureVector
        \For{$i = 1$ to number of cells}
            \State cell size $S_i$ = number of voxel inside cell $i \ \times$ voxel resolution
            \State $wdeg(v_i)=\frac{\sum_{j} w_{ij}}{deg(v_i)}$
            \State append $(S_i,deg(v_i),wdeg(v_i))$ to FeatureVector
        \EndFor
       \EndFunction
\end{algorithmic}
\end{algorithm}

After computing $\mathbf{f}_{track}^i$ for all nodes in two consecutive frames, we link two nodes from different frames based on the following similarity measurement $sim(i,j)$ defined as
\begin{equation}
\begin{aligned}
    sim(i,j) = \frac{|S_{1i}-S_{2j}|}{S_{1i}}+\frac{|deg_1(v_i)-deg_2(v_j)|}{deg_1(v_i)}+\\
    \begin{aligned}
    \frac{|\text{wdeg}_1(v_i)-\text{wdeg}_2(v_j)|}{\text{wdeg}_1(v_i)}
    \end{aligned}
\end{aligned}
\end{equation}
where $i$ and $j$ denote two nodes from two consecutive frames. 
We define $sim$ so that we can allow different units of entries in $\mathbf{f}_{\text{track}}^i$. 
We find $i^{*}$ and $j^{*}$ that minimizes $sim$. $i^{*}$ and $j^{*}$ are linked only when their $sim$ is below a set threshold value. In our experiments, the threshold we use is between 0.1 to 0.5.

\section{Dataset}
\subsection{Plasma-membrane Tagged Dataset}
Two 3D confocal image stack datasets of fluorescent-tagged plasma-membrane cells are used in this paper. In both datasets, only the plasma-membrane signal is used and is represented by voxels with high intensity values.  
The first dataset \cite{purdue} (Dataset 1) consists of a long-term time-lapse from \textit{A. thaliana's} leaf epidermal tissue that spans over a 12 hour period with a xy-resolution of 0.212$\mu m$ and 0.5$\mu m$ thick optical sections. There are 5 sequences of image stacks. Each sequence has 9-20 image stacks and each stack has 18 to 25 slices containing one layer of cells, and the dimensions of each slice is $512\times 512$.
Partial ground truth sub-cellular features are provided for this dataset.
Details of this dataset are described in Table \ref{tab:celldataset1}.

The second dataset (Dataset 2) contains cells in the shoot apical meristem of 6 \textit{Arabidopsis thaliana} \cite{data}. 
There are 6 image sequences.
Each image sequence has 20 image stacks.
In each image stack, there are 129 to 219 slices containing of 3 layers ($L$) of cells: outer layer ($L_1$), middle layer ($L_2$), and deep layer ($L_3$), and the dimension of each slice is $512\times 512$.
The available resolution of each image in x and y direction are 0.22$\mu m$ and in z is about 0.26$\mu m$. 
The ground truth voxel-wise cell labels are provided, and each cell has a unique label.
Each cell track also has a unique track ID.
Details of this dataset are described in Table \ref{tab:celldataset2}


\begin{table*}[ht]
\scriptsize
    \centering
    \caption{Single Layer Pavement Cell Dataset \cite{purdue}. It consists of a long-term time-lapse from \textit{A. thaliana's} leaf epidermal tissue that spans over a 12 hour period with a xy-resolution of 0.212$\mu m$ and 0.5$\mu m$ thick optical sections. The time step is two hours for sequence\#2 and is one hour for all other sequences. Anticlinal cell walls are partially annotated for all sequences. In addition to that, cells are partially annotated for sequence\#5 }
    \begin{tabular}{|c|c|c|}
    \hline
         Dataset& Number of Time Points&Image Stack Dimension\\
         \hline
         Sequence 1 &20&$512\times512\times20$\\
        \hline
        Sequence 2 &9&$512\times512\times18$\\
        \hline
        Sequence 3  &10&$512\times512\times30$\\
        \hline
        Sequence 4  &13& $512\times512\times21$\\
        \hline
        Sequence 5  &20&$512\times512\times25$\\
        \hline
    \end{tabular}
    \label{tab:celldataset1}
\end{table*}

\begin{table*}[ht]
\scriptsize
    \centering
    \caption{Multi Layer Pavement Cell Dataset \cite{data}. It contains three layers of cell walls in the shoot apical meristem of \textit{A. thaliana's} that spans over 80 hours with with a xy-resolution of 0.22$\mu m$ and 0.26$\mu m$ thick optical sections. The time step is 4 hours for all sequences and each sequence has 20 frames. Cells with track IDs are fully provided.}
    \begin{tabular}{|c|c|}
    \hline
         Dataset& Image Stack Dimension\\
         \hline
         Sequence 1& $512\times512\times134$ \\
        \hline
        Sequence 2&  $512\times512\times219$\\
        \hline
        Sequence 3&  $512\times512\times119$\\
        \hline
        Sequence 4&  $512\times512\times129$\\
        \hline
        Sequence 5&  $512\times512\times139$\\
        \hline
        Sequence 6& $512\times512\times134$\\
        \hline
    \end{tabular}
    \label{tab:celldataset2}
\end{table*}

\subsection{Cell Nuclei Dataset}
The 3D time-lapse video sequences of fluorescent nuclei microscopy image of \textit{C.elegans} developing embryo (Dataset 3). Each voxel size is $0.09 \times 0.09 \times 1.0 $ in microns. Time points were collected once per minute for five to six hours. There are two videos in the training set and two videos in the testing dataset. Details of this dataset are described in Table \ref{tab:nucleidata}. This dataset is used to evaluate our tracking algorithm performance.

\begin{table*}[ht]
\scriptsize
    \centering
    \caption{C.elegans Developing Embryo Nuclei Dataset \cite{ulman2017objective, 10.1093/bioinformatics/btu080}. The resolution of each image stack is $0.09\mu m\times0.09\mu m\times1.0\mu m$. Sequence 1 and 2 are training set which contains partial nuclei segmentation with track IDs for training. Sequence 3 and 4 are testing set so no annotations available.}
    \begin{tabular}{|c|c|c|c|}
        \hline
    Dataset & Time Step (min) & Number of Frames& Image Stack Dimension \\ \hline
        
        Sequence 1 & 1 & 250 & $512\times708\times35$  \\
        \hline
        Sequence 2 & 1.5 & 250 & $512\times712\times31$ \\
        \hline
        Sequence 3 & 1 & 190 & $512\times712\times31$ \\
        \hline 
        Sequence 4 & 1.5 & 140 & $512\times712\times31$ \\
        \hline
    \end{tabular}
    
    \label{tab:nucleidata}
\end{table*}
\section{Results}

\subsection{Segmentation}

\begin{figure*} [ht]
\centering
\begin{overpic}[width=0.16\textwidth]{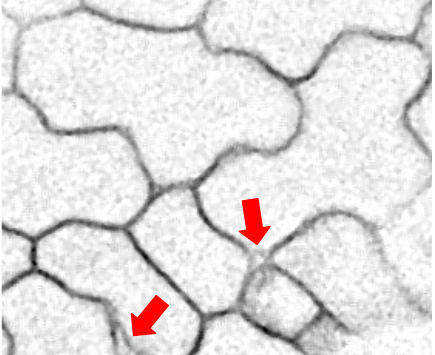}\put(5,4){A}\end{overpic}
\begin{overpic}[width=0.16\textwidth]{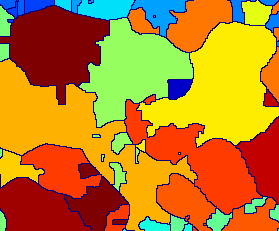}\put(5,4){B}\end{overpic}
\begin{overpic}[width=0.16\textwidth]{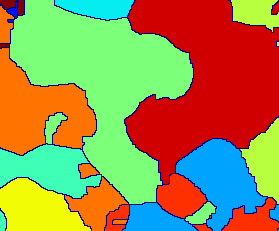}\put(5,4){C}\end{overpic}
\begin{overpic}[width=0.16\textwidth]{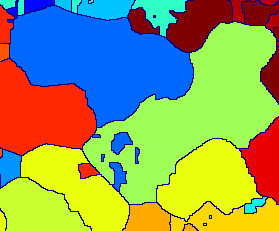}\put(5,4){D}\end{overpic}
\begin{overpic}[width=0.16\textwidth]{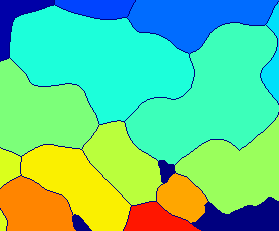}\put(5,4){E}\end{overpic}

\begin{overpic}[width=0.16\textwidth]{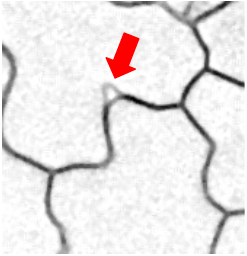}\put(5,4){A}\end{overpic}
\begin{overpic}[width=0.16\textwidth]{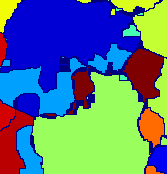}\put(5,4){B}\end{overpic}
\begin{overpic}[width=0.16\textwidth]{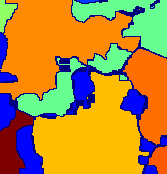}\put(5,4){C}\end{overpic}
\begin{overpic}[width=0.16\textwidth]{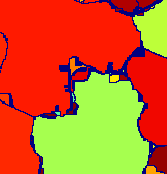}\put(5,4){D}\end{overpic}
\begin{overpic}[width=0.16\textwidth]{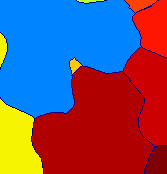}\put(5,4){E}\end{overpic}

\begin{overpic}[width=0.16\textwidth]{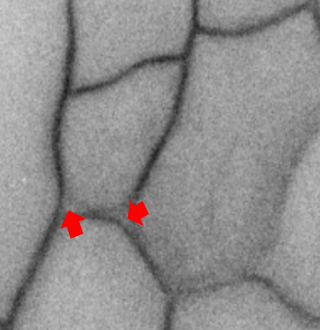}\put(5,4){A}\end{overpic}
\begin{overpic}[width=0.16\textwidth]{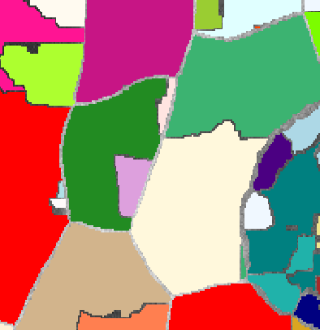}\put(5,4){B}\end{overpic}
\begin{overpic}[width=0.16\textwidth]{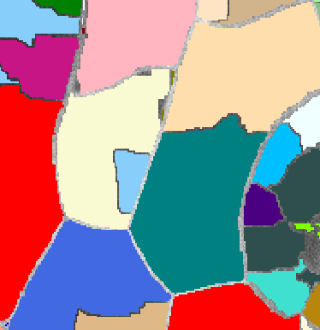}\put(5,4){C}\end{overpic}
\begin{overpic}[width=0.16\textwidth]{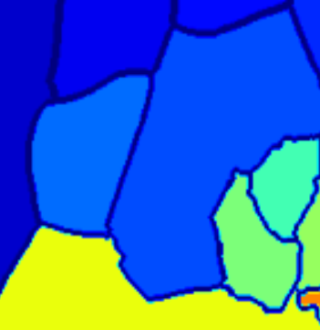}\put(5,4){D}\end{overpic}
\begin{overpic}[width=0.16\textwidth]{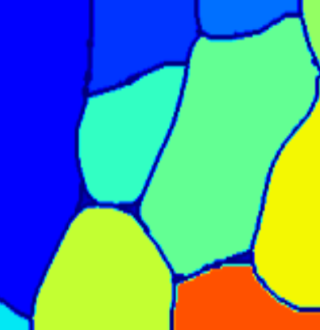}\put(5,4){E}\end{overpic}

\begin{overpic}[width=0.16\textwidth]{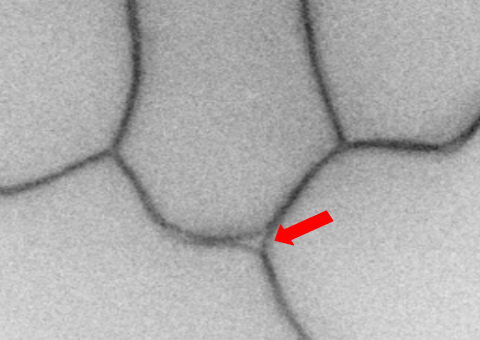}\put(5,4){A}\end{overpic}
\begin{overpic}[width=0.16\textwidth]{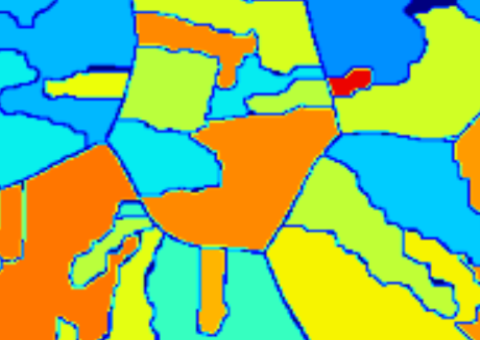}\put(5,4){B}\end{overpic}
\begin{overpic}[width=0.16\textwidth]{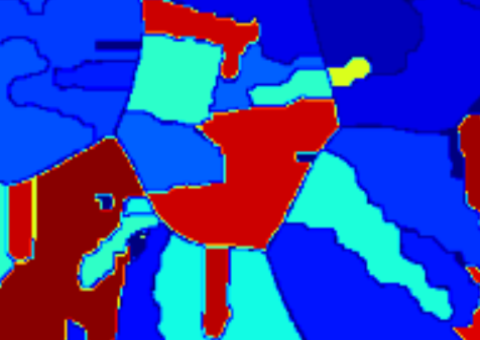}\put(5,4){C}\end{overpic}
\begin{overpic}[width=0.16\textwidth]{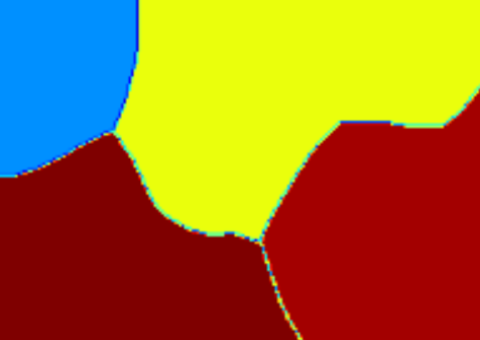}\put(5,4){D}\end{overpic}
\begin{overpic}[width=0.16\textwidth]{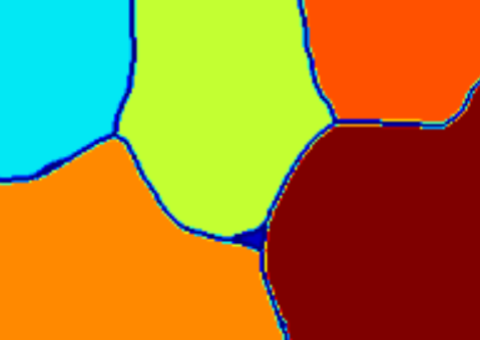}\put(5,4){E}\end{overpic}

\begin{overpic}[width=0.16\textwidth]{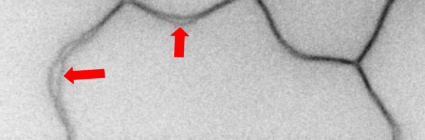}\put(5,4){A}\end{overpic}
\begin{overpic}[width=0.16\textwidth]{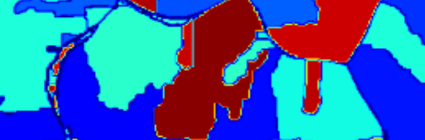}\put(5,4){B}\end{overpic}
\begin{overpic}[width=0.16\textwidth]{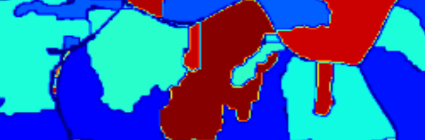}\put(5,4){C}\end{overpic}
\begin{overpic}[width=0.16\textwidth]{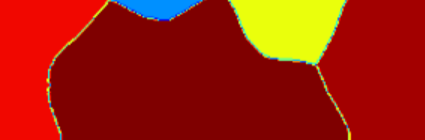}\put(5,4){D}\end{overpic}
\begin{overpic}[width=0.16\textwidth]{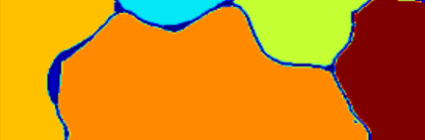}\put(5,4){E}\end{overpic}
\caption{The figure shows the segmentation results of the cell image with inter-cellular space or \textit{protrusion} indicated by a red arrow. (A) Inverted raw image in xy orientation, (B) MARS, (C) ACME, (D) supervoxel-based method, (E) proposed method.}
\label{fig:result2}
\end{figure*}

We apply our proposed method to the Dataset 1 for the purpose of identifying and analyzing cells based on the segmentation. 
The segmentation results of our proposed method and other state-of-the-art methods are shown in Fig.~\ref{fig:result2}. 
Our proposed method has visually better segmentation performance with closed cell surface and smooth boundary, and our method is able to identify the inter-cellular spaces and \textit{protrusions} in the 3D cell image stack. 
For Dataset 1, we do not have full cell annotations, so we only evaluate the cell counting accuracy on this dataset.

\begin{table*}[ht]
\scriptsize
    \centering
    \caption{Cell Counting Accuracy for Different Methods. For each time sequence, there is a fixed number of cells. Due to segmentation error, the algorithms can generate different number of cells for different time points of the sequence.In the table, we showed average and standard deviation of number of detection cells for the whole sequence}
    \begin{tabular}{|c|c|c|c|c|c|}
        \hline
    Sequence & Ground truth &ACME\cite{mosaliganti2012acme} & MARS \cite{mars}& Supervoxel Method \cite{supervoxelmerge} &Our method\\ 
        \hline
        
        Sequence 1 & 23 & 21.5 $\pm$3.2 & 25.5 $\pm$2.2 & 24 $\pm$ 1.1& \textbf{23.5 $\pm$ 0.9}  \\
        \hline
        Sequence 2 & 30 & 41.1 $\pm$3.1 & 35.1 $\pm$2.8 & 32 $\pm$ 2.1& \textbf{30.1 $\pm$ 0.8}  \\
        \hline
        Sequence 3 & 25 & 22.6 $\pm$2.1 & 27.5 $\pm$3.2 & 24 $\pm$ 1.5& \textbf{25 $\pm$ 0.5}  \\
        \hline 
        Sequence 4 & 18 & 18.8 $\pm$1.2 & 18.5 $\pm$1.2 & 18.2 $\pm$ 1.2& \textbf{18 $\pm$ 0.6}  \\
        \hline
        Sequence 5 & 28 & 31.5 $\pm$2.9 & 24.5 $\pm$2.3 & 26.2 $\pm$ 1.1& \textbf{27.8 $\pm$ 1}  \\
        \hline
    \end{tabular}
  
    \label{tab:cellcountingdata1}
\end{table*}

For each sequence, there are a fixed number of cells for all time points. Therefore, we want segmentation algorithms to generate average cell counting results close to ground truth counting numbers, and the variance of counting results for one sequence should be as small as possible. 
Details of the cell counting results are in Table \ref{tab:cellcountingdata1}.
Clearly, our method has the best cell counting performance. 

\begin{figure}[ht]
  \centering
  \centerline{\includegraphics[width=0.7\columnwidth]{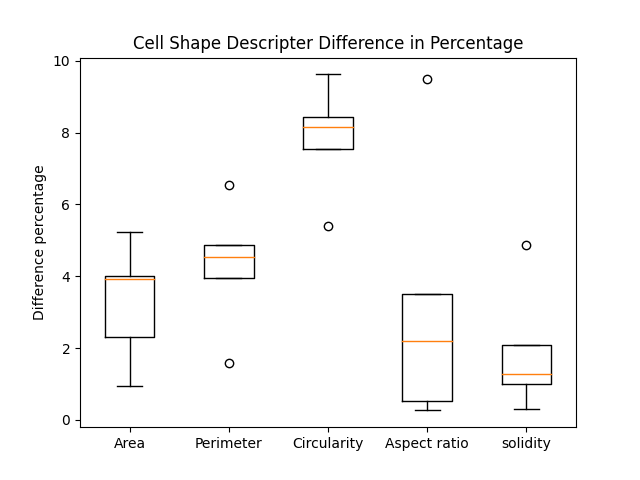}}
  \caption{3D segmentation evaluation using cell shape descripter including area, perimeter, circularity, aspect ratio, and solidity (ratio between cell area and its convex hull area). The difference is in terms of percentage.}
  \label{fig:shape evaluation}
\end{figure}

In order to verify if the output of the segmentation can be used for time lapse sequence analysis, we calculate basic cell shape information from the maximum area plane of the cells to compare with the expert annotations. The maximum area plane of a cell is the image plane which has the largest cell area across all z-slices. The shape information includes area, perimeter, circularity, and solidarity. Fig.~\ref{fig:shape evaluation} shows the comparison. Note that not all cells are annotated so that some cell comparisons are missed. The average shape difference is 4.5 percent and the largest shape difference is within 10 percent. 

Next, we apply our cell segmentation method on Dataset 2. 
Half of the $L_1$ layer of the training set is used to train the 3D U-Net, and the remaining $L_1$, $L_2$ and $L_3$ layers of the testing set are used to evaluate the segmentation performance of the proposed algorithm. 
To measure the boundary segmentation accuracy, for each labeled cell wall voxel in ground truth, we view it as the binary detection problem.
If there is a detected boundary voxel by any boundary detection algorithm within 5 voxels of a ground truth boundary voxel, then it is a correct detection.
If there is not any detected boundary voxel by algorithms within 5 voxels of a ground truth boundary voxel, then it is a miss detection.
If there is a detected boundary voxel by algorithms within 5 voxels of a voxel that is not ground truth boundary voxel, then it is a miss detection.
Based on this binary detection, we calculate the boundary segmentation precision, recall, and F-score.
Table~\ref{tab:l1} to \ref{tab:l3} shows the comparison of the final segmentation boundary result using our proposed method and other methods including ACME \cite{mosaliganti2012acme}, MARS \cite{mars} and a supervoxel-based algorithm \cite{supervoxelmerge} on $L_1$ to $L_3$ respectively. 
In terms of cell wall accuracy, our model shows at least 0.03 improvement in the F-score measure on average in terms of cell wall segmentation accuracy. 

It is noted that the average segmentation time of our proposed model is significantly shorter compared to the supervoxel-based method \cite{supervoxelmerge}.
Our proposed method takes approximately 0.8 seconds to segment one 512$\times$512 image slice on average, whereas supervoxel-based method takes approximately 6 seconds on a NVIDIA GTX Titan X with an Intel Xeon CPU E5-2696 v4 @ 2.20GHz. 
\begin{table}[ht]
\scriptsize
\centering
\caption{3D Segmentation Performance on $L_1$. If there is a detected boundary voxel by algorithms within 5 voxels of a ground truth boundary voxel, then it is a correct detection.
If there is not any detected boundary voxel by algorithms within 5 voxels of a ground truth boundary voxel, then it is a miss detection.
If there is a detected boundary voxel by algorithms within 5 voxels of a voxel that is not ground truth boundary voxel, then it is a miss detection. Same evaluation metric for $L_2$ and $L_3$}
\label{tab:l1}
\begin{tabular}{|c|c|c|c|}
    \hline
     Algorithm & Precision&Recall&F-Score \\
    \hline
     ACME \cite{mosaliganti2012acme}&0.805 &0.966 &0.878\\
     MARS \cite{mars}&0.910 &0.889 &0.899\\
     Supervoxel method \cite{supervoxelmerge}&\textbf{0.962} &0.932 &0.947\\
     our method&0.961 &\textbf{0.973} &\textbf{0.967}\\
     \hline
\end{tabular}
\centering
\caption{3D Segmentation Performance on $L_2$}
\label{tab:l2}
\begin{tabular}{|c|c|c|c|}
    \hline
     Algorithm & Precision&Recall&F-Score \\
    \hline
     ACME \cite{mosaliganti2012acme}&0.775 &\textbf{0.980} &0.866\\
     MARS \cite{mars}&0.921 &0.879 &0.900\\
     Supervoxel method \cite{supervoxelmerge} &0.910 &0.932 &0.921\\
     our method&\textbf{0.955} &0.971 &\textbf{0.963}\\
     \hline
\end{tabular}
\centering
\caption{3D Segmentation Performance on $L_3$}
\label{tab:l3}
\begin{tabular}{|c|c|c|c|}
    \hline
     Algorithm & Precision&Recall&F-Score \\
    \hline
     ACME \cite{mosaliganti2012acme}&0.745 &\textbf{0.976} &0.845\\
     MARS \cite{mars}&0.909 &0.879 &0.894\\
     Supervoxel method \cite{supervoxelmerge} &\textbf{0.982} &0.881 &0.929\\
     our method&0.955 &0.942 &\textbf{0.949}\\
     \hline

\end{tabular}
\end{table}

\subsection{Tracking and Feature Extraction}
\label{results}

\begin{figure}[ht] 
	\centering
	\begin{overpic}[width=0.216\textwidth]{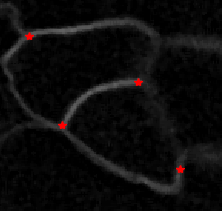}\put(5,3){}\textcolor{white}{A}\end{overpic}
	\begin{overpic}[width=0.196\textwidth]{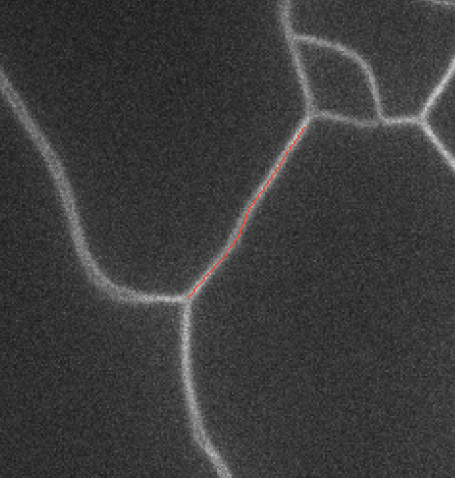}\put(5,3){}\textcolor{white}{B}\end{overpic}
    \caption{A: Extracted junctions of three cell walls, B: Extracted anticlinal wall segment.}
    \label{fig:shape_comparison}
\end{figure}

We apply our whole workflow on Dataset 1 to extract sub-cellular features like anticlinal wall segments and junctions of 3 cell walls. 
Qualitative results of the extracted sub-cellular features are shown in Fig.~\ref{fig:shape_comparison}.

\begin{figure}[ht] 
	\centering
	\begin{overpic}[width=0.216\textwidth]{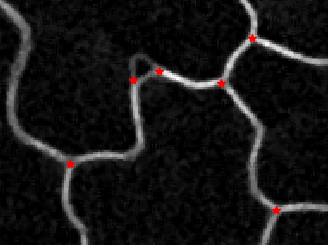}\put(90,2){}\textcolor{white}{A} \end{overpic}
	\begin{overpic}[width=0.216\textwidth]{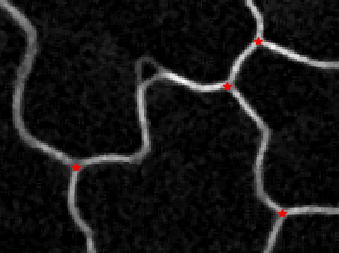}\put(90,2){}\textcolor{white}{B}\end{overpic}
    \caption{Example of computing 3 cell wall junctions from raw image (A), and (B) using our method.}
    \label{fig:junctioncomparison}
\end{figure}

The quantitative measurement of accuracy of junctions of 3 cell walls is also provided. 
We compare our results with 3D corner detection based method \cite{traditioncorner} on the raw image stack, and applying our 3 cell wall junction detection method using the segmentation image from other state-of-the-art methods \cite{mosaliganti2012acme,mars,supervoxelmerge}. The 3 cell wall junction detection results are shown in Table \ref{tab:junctions}. If 3 cell wall junctions are detected within 5 voxels of a ground truth 3 cell wall junction, it is a correct detection. Then we define false positive (FP), and false negative (FN) based on the binary detection of 3 cell walls junction. 
The error (E) is defined by the summation of FP and FN and normalized by total number of true 3 cell wall junctions. 
The results in the table \ref{tab:junctions} are average values across all image stacks.
From the table, we can see our method has the best 3 cell wall junction detection accuracy in terms of E.
Compared to the method that directly computes 3 cell wall junctions from raw image, our method has significantly better performance in terms of FP. 
This is because not all corner points are junctions of three cell walls. 
For example, corner detection based method gives false positive in the case shown in Fig.~\ref{fig:junctioncomparison}. Our graph based image feature extraction model not only uses low level image features but also some semantic information. 
 
\begin{table}[]
    \centering
    \caption{Quantitative Analysis on Error of Junctions of Three Cell Walls. FP is average number of false positive 3 cell wall detection for one 3D image stack. FN is average number of false negative 3 cell wall detection for one 3D image stack. E is the summation of FP and FN and normalized by total number of true 3 cell wall junctions}
    \begin{tabular}{|c|c|c|c|}
    \hline
         Algorithm& FP&FN&E \\
         \hline
         Corner Detection\cite{traditioncorner}& 6.23&1.98&0.15\\
         ACME \cite{mosaliganti2012acme}&10.33&2.11&0.23\\
         MARS \cite{mars}&12.54&\textbf{1.11}&0.26\\
         Supervoxel method \cite{supervoxelmerge}&3.27&2.68&0.11\\
         Our Method&\textbf{1.21}&1.77&\textbf{0.05}\\
    \hline
    \end{tabular}
    \label{tab:junctions}
\end{table}

The anticlinal wall segment is defined by two neighboring junctions of 3 cell walls are also computed. 
The partial annotation of such segments are provided.
We would like to note that such manual annotations are very labor intensive and it is impractical to annotate all anticlinal cell wall segments (see Fig \ref{fig:workflow}) even in a single 3D volume. The practical difficulties include lack of support for 3D visualization and annotation tools for tracing.
The ground truth segments were annotated by going through each slice in the image stack, finding the approximate slice where neighboring cell walls touch, and then tracing the segment in that single slice. Each segment in the ground truth is represented by a collection of coordinates of the segment in that image slice. Note that different segments can be on different slices. In contrast, each of our computed segments can span multiple Z slices, hence providing a more accurate 3D representation than is manually feasible. This also makes it challenging to compare the manual ground truth with the computed results.

We define a set of evaluation metrics to evaluate how accurate our extracted anticlinal wall segments are compared to their partial annotations. 
Given the ground truth segment and the corresponding computed segment, the following evaluation metrics are computed: 
\begin{enumerate}
    \item End-point Displacement error (EDE) in the end points of the two segments;
    \item Fr\'echet distance (FD) between the two segments. FD is a measure of shape similarity of two curves and it takes into account the location and ordering of points along the curves.
Mathematically, consider two curves $P$ and $Q$,  FD $F(P,Q)$ between them is defined in the following equation
\begin{equation}
\begin{aligned}
    F(P,Q) = \inf_{\alpha,\beta} \max_{t\in[0,1]}d(P(\alpha(t)),Q(\beta(t)))
\end{aligned}
\end{equation}
where d is the Euclidean distance, $\alpha:[0,1]\rightarrow[0,m]$, $\beta:[0,1]\rightarrow[0,n]$ are all monotone parameterization of $[0,m]$ and $[0,n]$ respectively. $m$ and $n$ are length of $P$ and $Q$.
    \item Length difference (LD), absolute difference in lengths between the two segments
    \item Percentage difference in length (DP), length difference normalized by ground truth length.
\end{enumerate}

\begin{table}[tbp]
    \centering
    \caption{Anticlinal cell wall segment evaluation on Dataset 1 using EDE, FD, LD, DP}
    \begin{tabular}{|c|c|c|c|c|}
    \hline
    Sequence & EDE & FD  & LD & DP \\ 
    \hline
       
        Sequence 1 & 2.80 & 4.00  & 2.32 &2.90 \\ 
        Sequence 2 & 6.40 & 8.40  & 4.19 &6.10 \\
        Sequence 3 & 3.05   &3.9 & 1.74&2.4 \\ 
        Sequence 4 &3.02 &3.70 &2.24 &2.3 \\
        Sequence 5 &2.33 &3.40 &2.11 &2.10 \\ \hline
    \end{tabular}

    \label{tab:segments}
    \vspace{-10pt}
\end{table}

Average EDE between the results using our method and the ground truth is 3.03 voxels, average FD is 3.7 voxels, average LD is 2.24 voxels, and average DP is 2.3 percent. 
Evaluation results of different time series are shown in Table \ref{tab:segments} and evaluation result for each segment is in the supplemental materials.

%
%

We also apply our tracking method on Dataset 2 and Dataset 3. 
Table \ref{tab:track_pavement} shows the quantitative comparison of our method with other state-of-the-art cell/nuclei tracking methods. The evaluation metric we use is tracking accuracy (TRA), proposed in \cite{trackevalue}. TRA measures how accurately each cell/nuclei is identified and followed in successive image stacks of the sequence. Ground truth tracking results and tracking results generated from algorithms are viewed as two acyclic oriented graphs and TRA measures the number of operations needed to modify one graph to another. More specifically, TRA is defined on Acyclic Oriented Graph Matching (AOGM) as 
\begin{align}
     \text{TRA} = 1 - \dfrac{\min(\text{AOGM}, \text{AOGM}_0)}{\text{AOGM}_0}
\end{align}
where AOGM$_0$ is the AOGM value required for creating the reference graph from scratch. TRA ranges between 0 to 1 (1 means perfect tracking).
Our method shows a rough 0.05 TRA measurement improvement on Dataset 2. 
To demonstrate the robustness of our tracking method, we also apply it on Dataset 3, a cell nuclei dataset, and achieve a TRA of 0.895 which is comparable to state-of-the-art tracking methods on IEEE ISBI CTC2020 cell tracking challenge. State-of-the-art methods (\cite{lofflerkit} and \cite{magnusson2014global}) are based on the traditional Viterbi cell tracking algorithm whose complexity is $\mathcal{O}(TM^{4})$ where $T$ is the length of the sequence and $M$ is the maximum number of cells/nuclei. In contrast, the complexity of our method is $\mathcal{O}(TM^{2})$. Sequence 1 and 2 are the training data released from the challenge and we run the state-of-the-art methods on the individual sequence to get TRA evaluation metric. Sequence 3 and 4 are testing data that is not published by the challenge and TRA values are given by the challenge organization.
\begin{table}[h]
    \centering
    \caption{Cell Tracking Performance on Dataset 2 and Dataset 3}
    \begin{tabular}{|c|c|c|c|}
    \hline
         Dataset 2& Viterbi Tracker\cite{track1}&Cell Proposal\cite{track3}&Our method \\
         \hline
         Sequence 1 &0.513 &0.492&\textbf{0.571}\\
         Sequence 2 &0.520 &0.512&\textbf{0.593}\\
         Sequence 3 &0.488 &0.532&\textbf{0.581}\\
         Sequence 4 &0.533 &0.498&\textbf{0.566}\\
         Sequence 5 &0.542 &0.525&\textbf{0.602}\\
         Sequence 6 &0.518 &0.542&\textbf{0.544}\\
    \hline
        Dataset 3 &KIT-Sch-GE \cite{lofflerkit}&KTH-SE \cite{magnusson2014global}&Our method \\
        \hline
        Sequence 1&0.903&\textbf{0.942}&0.931\\
        Sequence 2 &0.906&0.893&\textbf{0.912}\\
        Sequence 3 and 4 &0.886&\textbf{0.945}&0.895\\
    \hline
    \end{tabular}
    \label{tab:track_pavement}
\end{table}



\section{Conclusion}
In this paper, we present an end-to-end workflow for extracting quantitative information from 3D time-lapse imagery. The workflow includes 3D segmentation, tracking, and sub-cellular feature extraction. The 3D segmentation pipeline utilizes deep learning models with rotation equivariance. Then an adjacency graph is built for cell tracking and sub-cellular feature extraction. We demonstrate the performance of our model on multiple cell/nuclei datasets. In addition, we also curate a new pavement cell dataset with partial expert annotations that will be made available to researchers. 

The proposed segmentation method is implemented as a computational module in BisQue \cite{kvilekval2010bisque, latypov2019bisque}. 
The BisQue platform \href{https://bisque2.ece.ucsb.edu}{\texttt{bisque2.ece.ucsb.edu}} helps researchers organize, analyze, visualize, annotate, and share multi-dimensional, multimodal imaging data \cite{kvilekval2010bisque, latypov2019bisque}. Further, BisQue enables reproducible image analysis and provenance tracking of the computations on the data. The segmentation module in BisQue is referred to as CellECT2.0.

Users can run the CellECT2.0 module using the following steps: (1) Navigate to BisQue on their web browser and create an account, (2) Upload their own data in TIFF format or use suggested example dataset, 
(3) Select an uploaded TIFF image or use our example, (4) Select \texttt{Run} and the BisQue service will compute the segmentation results and display it in the browser. The runtime for a $512 \times 512 \times 18$ image is approximately one minute using a CPU node with a 24 core Xeon processor and 128GB of RAM. We provide screenshots of these steps in the supplemental materials.

\section{Acknowledgement}
This work was supported by the National Science Foundation award No.1715544 to DBS and BSM, and the National Science Foundation SSI award No.1664172 to BSM.

\section{Author Contributions}
JJ designed and carried out the overall computer vision method pipeline including segmentation, tracking, and feature extraction, helped implement software into Bisque and drafted manuscript. AK dockerized the segmentation code, integrated the module into BisQue, and helped manuscript preparation. SS helped get cell nuclei tracking results, and helped manuscript preparation. SB obtained membrane tagged cell dataset, manually segmented sub-cellular features, and helped prepare the manuscript. MG helped visualization of segmentation results on BisQue and optimized adjacency graph computation. DBS coordinated membrane tagged cell dataset
acquisition, helped analyzed the data, and reviewed manuscript. BSM coordinated the overall design, development and evaluations of the image processing methods, and
helped prepare the manuscript.

\section{Data Availability}
The code is available on \href{https://github.com/UCSB-VRL/Time-lapse3DCellAnalysis}{\texttt{GitHub}}.
Dataset 1, Dataset 2, and Dataset 3 analyzed during the paper are available in the repository, \href{https://bisque2.ece.ucsb.edu}{\texttt{Dataset 1}}, \href{https://www.repository.cam.ac.uk/handle/1810/262530?show=full}{\texttt{Dataset 2}}, and
 \href{http://celltrackingchallenge.net/3d-datasets/}{\texttt{Dataset 3}} separately.

%










\bibliographystyle{IEEEtran}
\bibliography{IEEEabrv,Bibliography}

\vfill


\end{document}


\title{Supplemental Materials}
\author{}
\maketitle

\section{Segmentation and sub-cellular feature extraction from BisQue}


\begin{figure}[ht]
    \includegraphics[scale=0.115]{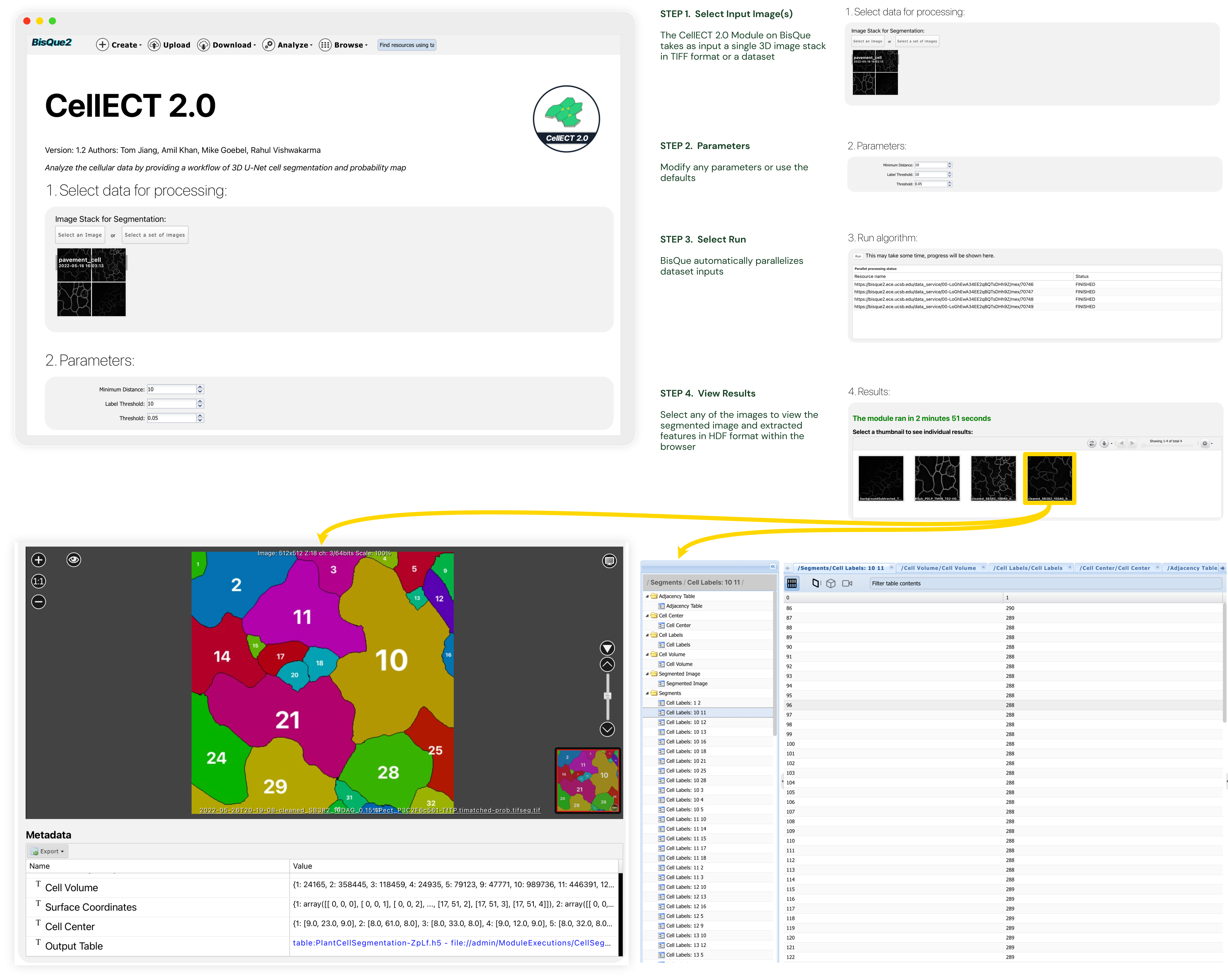}
    \caption{\textbf{CellECT2.0 Module in BisQue. } Here we have an example of four 3D stacked images from different time sequences of Dataset 1 in TIFF format as input to our module. BisQue will automatically parallel process the input images and return the results for each image. Each segmented cell is labeled with an integer in the segmentation image and each of the outputs have an associated HDF file containing the cellular and sub-cellular features, as summarized in Table \ref{tab:features}.}
    \label{fig:module}
\end{figure}

Cellular and sub-cellular features provided by BisQue are summarized in table \ref{tab:features}. 

\begin{table}[ht]
    \centering
    \caption{Cellular and Sub-Cellular Features Provided by BisQue}
    \begin{tabular}{|l|l|}
    \hline
        \textbf{Different features} & \textbf{Example Values or Explanation} \\
         \hline
         Cell Volume& number of voxels inside a cell, for example, 358,445\\
         \hline
         Neighboring (Adjacent) cells& for example, cells 1,3,11,14 are neighboring cells of cell 2\\ 
         \hline
         3D cell surface& list of coordinates of surface of points for one cell\\
         \hline
         Three cell wall junction points& for example, junction point of cell 2,3,11 is (12,207)\\
         \hline
         Cell Center& for example, center of cell 2 is (8,61,8)\\
         \hline
         Segments& list of coordinates of points along that segment\\
         \hline
    \end{tabular}
    \label{tab:features}
\end{table}

The full cellular and sub-cellular results for this image stack can be accessed \href{https://bisque2.ece.ucsb.edu/client_service/view?resource=https://bisque2.ece.ucsb.edu/data_service/00-oHh2Js78oyQHagKov6K9rk}{here}. 


\section{Segments evaluation}
Some examples of segments evaluation are summarized in Table \ref{tab:segments}. The evaluation metrics include: Euclidean error of start (end) points of segments, Fr\'echet Distance, and length difference of segments. 
In Table \ref{tab:segments}, the first column is segment ID. Segment ID includes two parts: Sequence ID and the segment ID within the sequence. Sequence ID starts with ``LGMTPM" and ends with a 2-digit number to differentiate different sequences. Segment ID contains information about which frame this segment comes from. For example, ``LGMTPM01 Seg03T28" means this segment comes from frame 28 of sequence 01. 

Detailed comparison results are in the \texttt{segments.xlsx} file.


\begin{table}[!h]
\caption{Segments evaluation results. The results include segments' end points location accuracy, segments length accuracy, and Frechet Distance for segments shape accuracy}
\begin{adjustbox}{width=\columnwidth,center}
\begin{tabular}{@{}llllllllllllllllll@{}}
\toprule
                  & \begin{tabular}[c]{@{}l@{}}detected start \\ point x\end{tabular} & \begin{tabular}[c]{@{}l@{}}detected start \\ point y\end{tabular} & \begin{tabular}[c]{@{}l@{}}detected start \\ point z\end{tabular} & \begin{tabular}[c]{@{}l@{}}detected end \\ point x\end{tabular} & \begin{tabular}[c]{@{}l@{}}detected end \\ point y\end{tabular} & \begin{tabular}[c]{@{}l@{}}detected end \\ point z\end{tabular} & \begin{tabular}[c]{@{}l@{}}GT start \\ point x\end{tabular} & \begin{tabular}[c]{@{}l@{}}GT start\\ point y\end{tabular} & \begin{tabular}[c]{@{}l@{}}GT end\\ point x\end{tabular} & \begin{tabular}[c]{@{}l@{}}GT end\\ point y\end{tabular} & \begin{tabular}[c]{@{}l@{}}euclidean distance\\ error start point\end{tabular} & \begin{tabular}[c]{@{}l@{}}euclidean distance\\ error end point\end{tabular} & \begin{tabular}[c]{@{}l@{}}Frechet \\ Distance\end{tabular} & \begin{tabular}[c]{@{}l@{}}GT\\ Length\end{tabular} & \begin{tabular}[c]{@{}l@{}}Detected \\ Length\end{tabular} & difference & \% change \\ \midrule
LGMTPM01 Seg03T28 & 391                                                               & 215                                                               & 3                                                                 & 485                                                             & 203                                                             & 3                                                               & 392.7                                                       & 216.2                                                      & 481.8                                                    & 203.3                                                    & 2.0                                                                            & 3.2                                                                          & 3.9                                                         & 95.1                                                & 93.4                                                       & 1.7        & 1.8       \\
LGMTPM01 Seg06T24 & 121                                                               & 144                                                               & 4                                                                 & 195                                                             & 142                                                             & 5                                                               & 124.9                                                       & 145.3                                                      & 196.0                                                    & 139.0                                                    & 4.1                                                                            & 3.2                                                                          & 4.1                                                         & 85.6                                                & 87.2                                                       & 1.6        & 1.9       \\
LGMTPM01 Seg02T01 & 169                                                               & 356                                                               & 2                                                                 & 234                                                             & 360                                                             & 4                                                               & 172.8                                                       & 358.3                                                      & 233.3                                                    & 360.3                                                    & 4.5                                                                            & 0.7                                                                          & 4.5                                                         & 66.6                                                & 70.3                                                       & 3.6        & 5.4       \\
LGMTPM01 Seg05T01 & 303                                                               & 274                                                               & 6                                                                 & 350                                                             & 320                                                             & 2                                                               & 303.5                                                       & 273.3                                                      & 346.8                                                    & 320.0                                                    & 0.9                                                                            & 3.3                                                                          & 3.3                                                         & 70.7                                                & 72.9                                                       & 2.2        & 3.1       \\
LGMTPM01 Seg08T01 & 200                                                               & 59                                                                & 1                                                                 & 299                                                             & 82                                                              & 2                                                               & 204.3                                                       & 59.8                                                       & 297.3                                                    & 81.8                                                     & 4.3                                                                            & 1.8                                                                          & 4.3                                                         & 100.6                                               & 98.1                                                       & 2.4        & 2.4       \\
LGMTPM02 Seg05T23 & 207                                                               & 138                                                               & 4                                                                 & 217                                                             & 77                                                              & 5                                                               & 210.2                                                       & 143.8                                                      & 213.3                                                    & 82.0                                                     & 6.6                                                                            & 6.2                                                                          & 8.4                                                         & 68.9                                                & 73.1                                                       & 4.2        & 6.1       \\
LGMTPM03 Seg02T32 & 178                                                               & 212                                                               & 4                                                                 & 246                                                             & 183                                                             & 4                                                               & 182.7                                                       & 212.3                                                      & 243.7                                                    & 182.7                                                    & 4.7                                                                            & 2.4                                                                          & 4.1                                                         & 77.8                                                & 80.3                                                       & 2.5        & 3.2       \\
LGMTPM04 Seg03T17 & 286                                                               & 327                                                               & 5                                                                 & 328                                                             & 392                                                             & 7                                                               & 289.7                                                       & 329.3                                                      & 325.2                                                    & 395.7                                                    & 4.3                                                                            & 4.6                                                                          & 4.7                                                         & 85.2                                                & 83.3                                                       & 2.0        & 2.3       \\
LGMTPM05 Seg03T01 & 45                                                                & 220                                                               & 3                                                                 & 145                                                             & 224                                                             & 5                                                               & 48.8                                                        & 219.3                                                      & 145.3                                                    & 220.3                                                    & 3.8                                                                            & 3.8                                                                          & 3.7                                                         & 106.5                                               & 108.1                                                      & 1.6        & 1.5       \\
LGMTPM05 Seg03T02 & 11                                                                & 198                                                               & 2                                                                 & 100                                                             & 200                                                             & 4                                                               & 13.5                                                        & 196.5                                                      & 101.5                                                    & 198.0                                                    & 2.9                                                                            & 2.5                                                                          & 3.0                                                         & 98.0                                                & 100.0                                                      & 2.0        & 2.0       \\
LGMTPM05 Seg03T01 & 12                                                                & 201                                                               & 3                                                                 & 100                                                             & 200                                                             & 5                                                               & 13.0                                                        & 202.5                                                      & 102.3                                                    & 203.5                                                    & 1.8                                                                            & 4.2                                                                          & 2.9                                                         & 99.3                                                & 97.0                                                       & 2.2        & 2.3       \\
LGMTPM05 Seg03T03 & 13                                                                & 200                                                               & 10                                                                & 103                                                             & 200                                                             & 8                                                               & 14.0                                                        & 199.8                                                      & 102.5                                                    & 201.0                                                    & 1.0                                                                            & 1.1                                                                          & 3.6                                                         & 98.5                                                & 102.1                                                      & 3.6        & 3.7       \\
LGMTPM05 Seg03T04 & 17                                                                & 208                                                               & 9                                                                 & 102                                                             & 211                                                             & 7                                                               & 15.5                                                        & 205.5                                                      & 104.5                                                    & 206.3                                                    & 2.9                                                                            & 5.4                                                                          & 3.7                                                         & 99.0                                                & 103.1                                                      & 4.1        & 4.1       \\
LGMTPM05 Seg03T05 & 16                                                                & 210                                                               & 8                                                                 & 105                                                             & 212                                                             & 8                                                               & 15.8                                                        & 209.3                                                      & 105.3                                                    & 210.3                                                    & 0.8                                                                            & 1.8                                                                          & 3.5                                                         & 99.5                                                & 101.2                                                      & 1.7        & 1.7       \\
LGMTPM05 Seg03T06 & 18                                                                & 184                                                               & 6                                                                 & 105                                                             & 185                                                             & 8                                                               & 17.0                                                        & 182.5                                                      & 106.8                                                    & 184.0                                                    & 1.8                                                                            & 2.0                                                                          & 3.2                                                         & 95.8                                                & 98.1                                                       & 2.3        & 2.4       \\
LGMTPM05 Seg03T07 & 19                                                                & 216                                                               & 7                                                                 & 104                                                             & 218                                                             & 9                                                               & 17.0                                                        & 215.8                                                      & 106.8                                                    & 216.5                                                    & 2.0                                                                            & 3.1                                                                          & 3.3                                                         & 99.8                                                & 101.2                                                      & 1.5        & 1.5       \\
LGMTPM05 Seg03T08 & 18                                                                & 221                                                               & 6                                                                 & 105                                                             & 221                                                             & 8                                                               & 17.0                                                        & 220.0                                                      & 106.5                                                    & 220.8                                                    & 1.4                                                                            & 1.5                                                                          & 3.6                                                         & 99.6                                                & 102.3                                                      & 2.7        & 2.7       \\
LGMTPM05 Seg03T09 & 18                                                                & 200                                                               & 6                                                                 & 110                                                             & 201                                                             & 7                                                               & 19.3                                                        & 199.0                                                      & 109.5                                                    & 199.8                                                    & 1.6                                                                            & 1.3                                                                          & 3.1                                                         & 100.3                                               & 101.3                                                      & 1.0        & 1.0       \\
LGMTPM05 Seg03T10 & 24                                                                & 192                                                               & 5                                                                 & 110                                                             & 195                                                             & 7                                                               & 22.0                                                        & 193.5                                                      & 112.3                                                    & 194.0                                                    & 2.5                                                                            & 2.5                                                                          & 3.3                                                         & 100.3                                               & 102.3                                                      & 2.1        & 2.0       \\
LGMTPM05 Seg03T11 & 23                                                                & 197                                                               & 6                                                                 & 115                                                             & 200                                                             & 9                                                               & 22.0                                                        & 196.9                                                      & 113.3                                                    & 197.8                                                    & 1.0                                                                            & 2.9                                                                          & 3.4                                                         & 101.3                                               & 101.2                                                      & 0.1        & 0.1       \\
LGMTPM05 Seg03T12 & 25                                                                & 200                                                               & 11                                                                & 117                                                             & 199                                                             & 9                                                               & 25.0                                                        & 198.5                                                      & 115.3                                                    & 198.3                                                    & 1.5                                                                            & 1.8                                                                          & 3.3                                                         & 100.3                                               & 101.4                                                      & 1.0        & 1.0       \\
LGMTPM05 Seg03T13 & 25                                                                & 201                                                               & 10                                                                & 114                                                             & 198                                                             & 8                                                               & 24.3                                                        & 197.3                                                      & 116.3                                                    & 200.5                                                    & 3.8                                                                            & 3.4                                                                          & 3.2                                                         & 102.1                                               & 104.3                                                      & 2.2        & 2.2       \\
LGMTPM05 Seg03T14 & 27                                                                & 200                                                               & 9                                                                 & 118                                                             & 200                                                             & 7                                                               & 25.5                                                        & 198.5                                                      & 118.0                                                    & 200.3                                                    & 2.1                                                                            & 0.3                                                                          & 3.6                                                         & 102.5                                               & 103.3                                                      & 0.7        & 0.7       \\
LGMTPM05 Seg03T15 & 26                                                                & 199                                                               & 7                                                                 & 119                                                             & 202                                                             & 7                                                               & 29.0                                                        & 197.8                                                      & 122.5                                                    & 198.3                                                    & 3.3                                                                            & 5.1                                                                          & 3.2                                                         & 103.5                                               & 101.5                                                      & 2.0        & 2.0       \\
LGMTPM05 Seg03T16 & 29                                                                & 195                                                               & 11                                                                & 122                                                             & 199                                                             & 12                                                              & 31.3                                                        & 196.3                                                      & 125.0                                                    & 197.8                                                    & 2.6                                                                            & 3.3                                                                          & 3.4                                                         & 103.8                                               & 100.3                                                      & 3.5        & 3.3       \\
LGMTPM05 Seg03T17 & 32                                                                & 220                                                               & 10                                                                & 126                                                             & 220                                                             & 11                                                              & 33.8                                                        & 221.0                                                      & 127.0                                                    & 222.8                                                    & 2.0                                                                            & 2.9                                                                          & 3.5                                                         & 103.3                                               & 101.3                                                      & 2.0        & 1.9       \\
LGMTPM05 Seg03T18 & 16                                                                & 230                                                               & 9                                                                 & 113                                                             & 230                                                             & 10                                                              & 17.0                                                        & 229.3                                                      & 111.8                                                    & 230.8                                                    & 1.3                                                                            & 1.5                                                                          & 3.7                                                         & 104.8                                               & 106.4                                                      & 1.6        & 1.5       \\
LGMTPM05 Seg03T19 & 35                                                                & 231                                                               & 8                                                                 & 130                                                             & 232                                                             & 9                                                               & 34.8                                                        & 230.0                                                      & 129.3                                                    & 230.8                                                    & 1.0                                                                            & 1.5                                                                          & 3.6                                                         & 104.5                                               & 107.9                                                      & 3.4        & 3.2       \\
LGMTPM05 Seg03T20 & 37                                                                & 220                                                               & 9                                                                 & 130                                                             & 224                                                             & 10                                                              & 37.5                                                        & 221.3                                                      & 132.3                                                    & 222.3                                                    & 1.3                                                                            & 2.9                                                                          & 3.9                                                         & 104.8                                               & 106.8                                                      & 2.0        & 1.9       \\
LGMTPM05 Seg03T21 & 31                                                                & 226                                                               & 8                                                                 & 129                                                             & 230                                                             & 8                                                               & 30.5                                                        & 225.5                                                      & 126.5                                                    & 227.0                                                    & 0.7                                                                            & 3.9                                                                          & 3.1                                                         & 106.0                                               & 101.3                                                      & 4.7        & 4.4       \\
LGMTPM05 Seg03T22 & 47                                                                & 220                                                               & 9                                                                 & 144                                                             & 223                                                             & 10                                                              & 48.8                                                        & 219.3                                                      & 145.3                                                    & 220.3                                                    & 1.9                                                                            & 3.0                                                                          & 4.0                                                         & 106.5                                               & 105.9                                                      & 0.6        & 0.6       \\
LGMTPM04 Seg01T01 & 365                                                               & 360                                                               & 12                                                                & 433                                                             & 425                                                             & 13                                                              & 366.0                                                       & 358.0                                                      & 429.3                                                    & 424.0                                                    & 2.2                                                                            & 3.9                                                                          & 3.1                                                         & 101.4                                               & 99.3                                                       & 2.2        & 2.1       \\
LGMTPM04 Seg01T02 & 366                                                               & 362                                                               & 13                                                                & 430                                                             & 427                                                             & 14                                                              & 365.8                                                       & 362.5                                                      & 429.0                                                    & 429.5                                                    & 0.6                                                                            & 2.7                                                                          & 3.2                                                         & 102.1                                               & 104.0                                                      & 1.8        & 1.8       \\
LGMTPM04 Seg01T03 & 365                                                               & 370                                                               & 12                                                                & 428                                                             & 436                                                             & 12                                                              & 368.0                                                       & 371.0                                                      & 431.3                                                    & 438.0                                                    & 3.2                                                                            & 3.8                                                                          & 3.2                                                         & 102.1                                               & 104.7                                                      & 2.5        & 2.5       \\
LGMTPM04 Seg01T04 & 367                                                               & 378                                                               & 11                                                                & 429                                                             & 449                                                             & 11                                                              & 367.3                                                       & 379.3                                                      & 431.5                                                    & 445.3                                                    & 1.3                                                                            & 4.5                                                                          & 4.0                                                         & 102.1                                               & 101.2                                                      & 0.9        & 0.9       \\
LGMTPM04 Seg01T05 & 365                                                               & 381                                                               & 12                                                                & 435                                                             & 448                                                             & 13                                                              & 367.8                                                       & 383.3                                                      & 431.3                                                    & 450.5                                                    & 3.6                                                                            & 4.5                                                                          & 4.2                                                         & 103.5                                               & 100.9                                                      & 2.6        & 2.5       \\
LGMTPM04 Seg01T06 & 363                                                               & 375                                                               & 14                                                                & 429                                                             & 453                                                             & 15                                                              & 368.3                                                       & 380.0                                                      & 433.0                                                    & 450.0                                                    & 7.3                                                                            & 5.0                                                                          & 3.6                                                         & 105.4                                               & 102.3                                                      & 3.0        & 2.9       \\
LGMTPM04 Seg01T07 & 369                                                               & 385                                                               & 11                                                                & 433                                                             & 450                                                             & 12                                                              & 369.3                                                       & 385.5                                                      & 430.8                                                    & 450.3                                                    & 0.6                                                                            & 2.3                                                                          & 3.4                                                         & 99.3                                                & 98.5                                                       & 0.8        & 0.8       \\
LGMTPM04 Seg01T08 & 370                                                               & 381                                                               & 11                                                                & 432                                                             & 455                                                             & 11                                                              & 370.3                                                       & 385.3                                                      & 432.8                                                    & 451.1                                                    & 4.3                                                                            & 4.0                                                                          & 3.1                                                         & 100.7                                               & 103.7                                                      & 3.0        & 3.0       \\
LGMTPM04 Seg01T09 & 371                                                               & 388                                                               & 10                                                                & 431                                                             & 452                                                             & 10                                                              & 371.0                                                       & 389.0                                                      & 433.0                                                    & 450.0                                                    & 1.0                                                                            & 2.8                                                                          & 3.9                                                         & 98.9                                                & 96.2                                                       & 2.6        & 2.7       \\
LGMTPM04 Seg01T10 & 373                                                               & 389                                                               & 9                                                                 & 433                                                             & 450                                                             & 9                                                               & 375.0                                                       & 389.3                                                      & 430.3                                                    & 451.0                                                    & 2.0                                                                            & 2.9                                                                          & 3.2                                                         & 92.8                                                & 95.0                                                       & 2.2        & 2.4       \\
LGMTPM04 Seg01T11 & 378                                                               & 389                                                               & 11                                                                & 433                                                             & 451                                                             & 9                                                               & 376.3                                                       & 389.3                                                      & 430.3                                                    & 452.3                                                    & 1.8                                                                            & 3.1                                                                          & 3.6                                                         & 93.0                                                & 90.0                                                       & 3.0        & 3.2       \\
LGMTPM04 Seg01T12 & 375                                                               & 389                                                               & 9                                                                 & 432                                                             & 455                                                             & 9                                                               & 377.8                                                       & 389.0                                                      & 430.3                                                    & 451.0                                                    & 2.8                                                                            & 4.4                                                                          & 4.9                                                         & 91.2                                                & 96.1                                                       & 4.9        & 5.3       \\
LGMTPM04 Seg01T13 & 366                                                               & 389                                                               & 8                                                                 & 443                                                             & 460                                                             & 9                                                               & 370.3                                                       & 391.3                                                      & 440.8                                                    & 458.3                                                    & 4.9                                                                            & 2.9                                                                          & 4.6                                                         & 107.1                                               & 103.2                                                      & 3.9        & 3.6       \\
LGMTPM04 Seg01T14 & 370                                                               & 390                                                               & 10                                                                & 439                                                             & 454                                                             & 10                                                              & 372.5                                                       & 392.3                                                      & 438.5                                                    & 451.0                                                    & 3.4                                                                            & 3.0                                                                          & 4.8                                                         & 98.3                                                & 96.1                                                       & 2.2        & 2.2       \\
LGMTPM04 Seg01T15 & 373                                                               & 386                                                               & 9                                                                 & 438                                                             & 453                                                             & 9                                                               & 374.3                                                       & 387.5                                                      & 436.0                                                    & 451.0                                                    & 2.0                                                                            & 2.8                                                                          & 3.2                                                         & 98.6                                                & 101.0                                                      & 2.4        & 2.5       \\
LGMTPM04 Seg01T16 & 372                                                               & 391                                                               & 9                                                                 & 430                                                             & 452                                                             & 11                                                              & 375.3                                                       & 389.5                                                      & 430.3                                                    & 451.0                                                    & 3.7                                                                            & 1.0                                                                          & 3.3                                                         & 92.5                                                & 93.9                                                       & 1.4        & 1.5       \\
LGMTPM04 Seg01T17 & 373                                                               & 394                                                               & 9                                                                 & 436                                                             & 452                                                             & 9                                                               & 375.0                                                       & 392.3                                                      & 435.0                                                    & 451.0                                                    & 2.7                                                                            & 1.4                                                                          & 3.2                                                         & 94.0                                                & 95.2                                                       & 1.3        & 1.4       \\
LGMTPM04 Seg01T18 & 378                                                               & 389                                                               & 9                                                                 & 435                                                             & 450                                                             & 9                                                               & 375.3                                                       & 386.0                                                      & 430.3                                                    & 451.0                                                    & 4.1                                                                            & 4.9                                                                          & 3.5                                                         & 95.2                                                & 92.0                                                       & 3.1        & 3.3       \\
LGMTPM04 Seg01T19 & 379                                                               & 389                                                               & 9                                                                 & 437                                                             & 455                                                             & 9                                                               & 377.5                                                       & 385.8                                                      & 435.3                                                    & 451.0                                                    & 3.6                                                                            & 4.4                                                                          & 3.7                                                         & 95.1                                                & 98.1                                                       & 3.0        & 3.1       \\
LGMTPM04 Seg01T20 & 378                                                               & 389                                                               & 9                                                                 & 438                                                             & 453                                                             & 9                                                               & 377.9                                                       & 388.0                                                      & 436.3                                                    & 451.0                                                    & 1.0                                                                            & 2.7                                                                          & 3.9                                                         & 95.9                                                & 96.1                                                       & 0.2        & 0.2       \\
LGMTPM04 Seg01T21 & 377                                                               & 389                                                               & 9                                                                 & 434                                                             & 452                                                             & 9                                                               & 374.3                                                       & 389.0                                                      & 431.0                                                    & 451.0                                                    & 2.7                                                                            & 3.2                                                                          & 4.1                                                         & 94.0                                                & 92.2                                                       & 1.8        & 1.9       \\
LGMTPM04 Seg01T22 & 370                                                               & 387                                                               & 9                                                                 & 439                                                             & 448                                                             & 9                                                               & 369.3                                                       & 386.0                                                      & 440.3                                                    & 451.0                                                    & 1.3                                                                            & 3.3                                                                          & 5.1                                                         & 106.3                                               & 102.0                                                      & 4.3        & 4.0       \\
LGMTPM04 Seg01T23 & 376                                                               & 389                                                               & 9                                                                 & 433                                                             & 449                                                             & 9                                                               & 375.1                                                       & 388.3                                                      & 430.3                                                    & 451.0                                                    & 1.2                                                                            & 3.4                                                                          & 3.2                                                         & 93.3                                                & 92.6                                                       & 0.7        & 0.8       \\
LGMTPM04 Seg01T24 & 372                                                               & 389                                                               & 9                                                                 & 433                                                             & 449                                                             & 9                                                               & 375.3                                                       & 389.3                                                      & 430.3                                                    & 451.0                                                    & 3.3                                                                            & 3.4                                                                          & 4.0                                                         & 92.6                                                & 92.7                                                       & 0.1        & 0.1       \\
LGMTPM03 Seg01T01 & 210                                                               & 371                                                               & 8                                                                 & 243                                                             & 417                                                             & 10                                                              & 212.5                                                       & 369.7                                                      & 243.0                                                    & 419.5                                                    & 2.8                                                                            & 2.5                                                                          & 4.2                                                         & 72.4                                                & 73.3                                                       & 0.8        & 1.2       \\
LGMTPM03 Seg01T02 & 215                                                               & 372                                                               & 9                                                                 & 245                                                             & 419                                                             & 11                                                              & 213.3                                                       & 370.0                                                      & 246.5                                                    & 421.5                                                    & 2.6                                                                            & 2.9                                                                          & 3.9                                                         & 71.3                                                & 73.6                                                       & 2.3        & 3.2       \\
LGMTPM03 Seg01T03 & 216                                                               & 375                                                               & 9                                                                 & 249                                                             & 420                                                             & 10                                                              & 214.5                                                       & 371.3                                                      & 248.0                                                    & 422.8                                                    & 4.0                                                                            & 2.9                                                                          & 3.6                                                         & 71.3                                                & 70.0                                                       & 1.3        & 1.9       \\ \bottomrule
\end{tabular}
\end{adjustbox}
\end{table}
\label{tab:segments}